\documentclass[11pt,a4paper]{article}
\usepackage[hyperref]{acl2021}
\usepackage{times}
\usepackage{latexsym}
\usepackage{amssymb}

\usepackage{microtype}

\usepackage{graphicx} 
\graphicspath{{assets/}} 
\usepackage{upquote}
\usepackage{textcomp}
\usepackage{amsmath}
\usepackage{arydshln}
\usepackage{enumitem}
\usepackage{hyperref}

\makeatletter
\newcommand\footnoteref[1]{\protected@xdef\@thefnmark{\ref{#1}}\@footnotemark}
\makeatother

\aclfinalcopy 
\def\aclpaperid{330} 

\title{LenAtten: An Effective Length Controlling Unit For Text Summarization}

\author{
Zhongyi Yu\Thanks{\hskip 0.5em Equal Contribution}\protect\phantom{\footnotesize 1}\textsuperscript{,}$^{1}$,\hskip 1em
{\bf Zhenghao Wu\footnotemark[1]\protect\phantom{\footnotesize 1}\textsuperscript{,}$^{1}$},\hskip 1em
{\bf Hao Zheng$^1$,}\\
{\bf Zhe XuanYuan$^2$\textsuperscript{,}$^{3}$,}\hskip 0.6em{\bf Jefferson Fong$^1$,}\hskip 1.3em{\bf Weifeng Su\Thanks{\hskip 0.5em Corresponding author}\protect\phantom{\footnotesize 1}\textsuperscript{,}$^{1}$\textsuperscript{,}$^{3}$} \\
  $^1$Computer Science and Technology Programme, $^2$Data Science Programme\\
  Division of Science and Technology\\
  $^3$Guangdong Key Lab of AI and Multi-Modal Data Processing\\BNU-HKBU United International College, Zhuhai, China \\
  \texttt{\{zhongyicst,zhecwu,zhslzwd97\}@gmail.com}\\
  \texttt{\{zhexuanyuan,jeffersonfong,wfsu\}@uic.edu.cn}\\
  }

\date{}

\begin{document}
\maketitle
\begin{abstract}
Fixed length summarization aims at generating summaries with a preset number of words or characters. Most recent researches incorporate length information with word embeddings as the input to the recurrent decoding unit, causing a compromise between length controllability and summary quality. In this work, we present an effective length controlling unit Length Attention (LenAtten) to break this trade-off. Experimental results show that LenAtten not only brings improvements in length controllability and ROGUE scores but also has great generalization ability. In the task of generating a summary with the target length, our model is 732 times better than the best-performing length controllable summarizer in length controllability on the CNN/Daily Mail dataset. \footnote{Code are publicly available at: \url{https://github.com/X-AISIG/LenAtten}} 

\end{abstract}

\section{Introduction}

Automatic text summarization aims at generating a short and coherent summary from one or multiple documents while preserving the main ideas of the original documents. Building upon the conventional summarization task, fixed length text summarization (FLS) demands extra focus on controlling the length of output summaries. Specifically, it requires generating summaries with a preset number of characters or words.

\begin{table}[!th]
\centering
\small
\resizebox{0.48\textwidth}{!}{%
\begin{tabular}{llcc}
\hline
\multicolumn{4}{l}{\textit{\textbf{Source document}}} \\ \hline
\multicolumn{4}{l}{\begin{tabular}{@{}l@{}}egyptian president hosni mubarak arrived here friday morning to\\ discuss the latest developments of iraqi crisis with his turkish\\ counterpart suleyman demirel .\end{tabular}} \\ \hline
\multicolumn{4}{l}{\textit{\textbf{Reference summary}}} \\ \hline
\multicolumn{4}{l}{\begin{tabular}{@{}l@{}}egyptian president to discuss iraqi crisis with turkish counterpart\end{tabular}} \\ \hline
\textit{\textbf{Model}} & \textit{\textbf{Summary}} \\ \hline
\textsc{Paulus} & \begin{tabular}{@{}l@{}}egyptian president arrives in ankara				\end{tabular} \\\hdashline
\textsc{Paulus}+LA2 (GT) & \begin{tabular}{@{}l@{}}mubarak arrives in ankara for talks on\\ iraqi crisis with turkish pm\end{tabular}  \\ \hline
\textsc{Paulus}+LA2 (30) & \begin{tabular}{@{}l@{}}egyptian president arrives in ankara\end{tabular}  \\ \hline
\textsc{Paulus}+LA2 (50) & \begin{tabular}{@{}l@{}}egyptian president arrives in ankara for \\talks on iraq crisis\end{tabular}\\ \hline
\textsc{Paulus}+LA2 (70) & \begin{tabular}{@{}l@{}}mubarak arrives in ankara for talks on \\iraqi crisis with turkish president demirel				\end{tabular} \\ \hline
\end{tabular}%
}
\caption{Output examples from the proposed method Length Attention (LA) on the Annotated English Gigaword dataset. Numbers in the parentheses represent different desired lengths. (GT) means the desired length is equal to the number of characters in the reference summary. \textsc{Paulus} \citep{DBLP:conf/iclr/PaulusXS18}.}
\label{tab:Intro_FLS_generation_examples}
\end{table}

FLS is a rising research topic required in many scenarios. For example, in order to get universal user experiences on multiple platforms and devices, titles and abstracts for news articles are expected to have different numbers of characters. Instead of manually rewriting summaries, FLS can automatically generate required summaries by simply inputting the desired output length. Besides, FLS can help news editors to reduce post-editing time \citep{makino-etal-2019-global} and further improve summary quality \citep{liu-etal-2018-controlling,makino-etal-2019-global}. Last but not least, as shown in Table \ref{tab:Intro_FLS_generation_examples}, with FLS, users can get customizable summaries by setting different desired lengths.

Despite the benefits that could be brought, previous studies on FLS are very limited. Recent researches in FLS apply length information to either (i) the decoder \citep{kikuchi-etal-2016-controlling,liu-etal-2018-controlling,takase-okazaki-2019-positional} or (ii) the optimization objective function \citep{makino-etal-2019-global}. Though these systems are promising, they have to make a compromise between length controllability and summary quality. \citet{kikuchi-etal-2016-controlling,makino-etal-2019-global} can generate high-quality summaries, but perform inadequately at controlling length. \citet{liu-etal-2018-controlling,takase-okazaki-2019-positional} control the output length accurately, but these models suffer from producing summaries with low ROUGE scores.

In this paper, we present an effective length controlling unit, Length Attention (LenAtten). With LenAtten, summarizers can generate high-quality summaries with a preset number of characters, successfully breaks the trade-off between length controllability and summary quality.

Our contributions in this work are as follows: (1) A novel length controlling unit with great generalization capability is proposed to make summarizers generate high-quality summaries with a preset number of characters. (2) Experimental results show that LenAtten can break the trade-off between length controllability and summary quality. The length controllability of the proposed method is the new state-of-the-art on the examined datasets, to our knowledge.

\section{Related Work}
There are two types of approaches for text summarization: the extractive approach and the abstractive approach. Extractive approaches generate summaries by extracting words or sentences from the original text \citep{dorr-etal-2003-hedge,nallapati2016summarunner-AAAI1714636,liu-lapata-2019-text,zhong-etal-2020-extractive}, while abstractive approaches produce novel words or phrases \citep{rush-etal-2015-neural,chopra-etal-2016-abstractive,DBLP:conf/conll/NallapatiZSGX16,gu-etal-2016-incorporating,see-etal-2017-get,fan-etal-2018-controllable,liu-lapata-2019-text}.

Derived from the works in general text summarization, two approaches have been developed for FLS: (1) Incorporating length information into the decoder. LenInit proposed in \citet{kikuchi-etal-2016-controlling} introduced length information into the initialization stage of a LSTM decoder. \citet{liu-etal-2018-controlling} follows a similar approach as LenInit, but it's based on a CNN sequence-to-sequence architecture. Other studies exploit length information in each decoding step. LenEmb introduced in \citet{kikuchi-etal-2016-controlling} generates a learnable embedding for each target length, and uses it as an additional input to its decoder. \citet{takase-okazaki-2019-positional} extended Transformer's sinusoidal positional encoding \citep{DBLP:conf/nips/VaswaniSPUJGKP17} to make summarizers take account of stepwise remaining length during prediction. (2) Leveraging length information in global optimization methods. \citet{makino-etal-2019-global} proposed a global optimization method named GOLC. GOLC incorporates length information with the minimum risk training (MRT) optimization method.

\section{Our Approach: Length Attention}
\begin{figure}[htbp]
  \centering
  \includegraphics[width=0.9\linewidth]{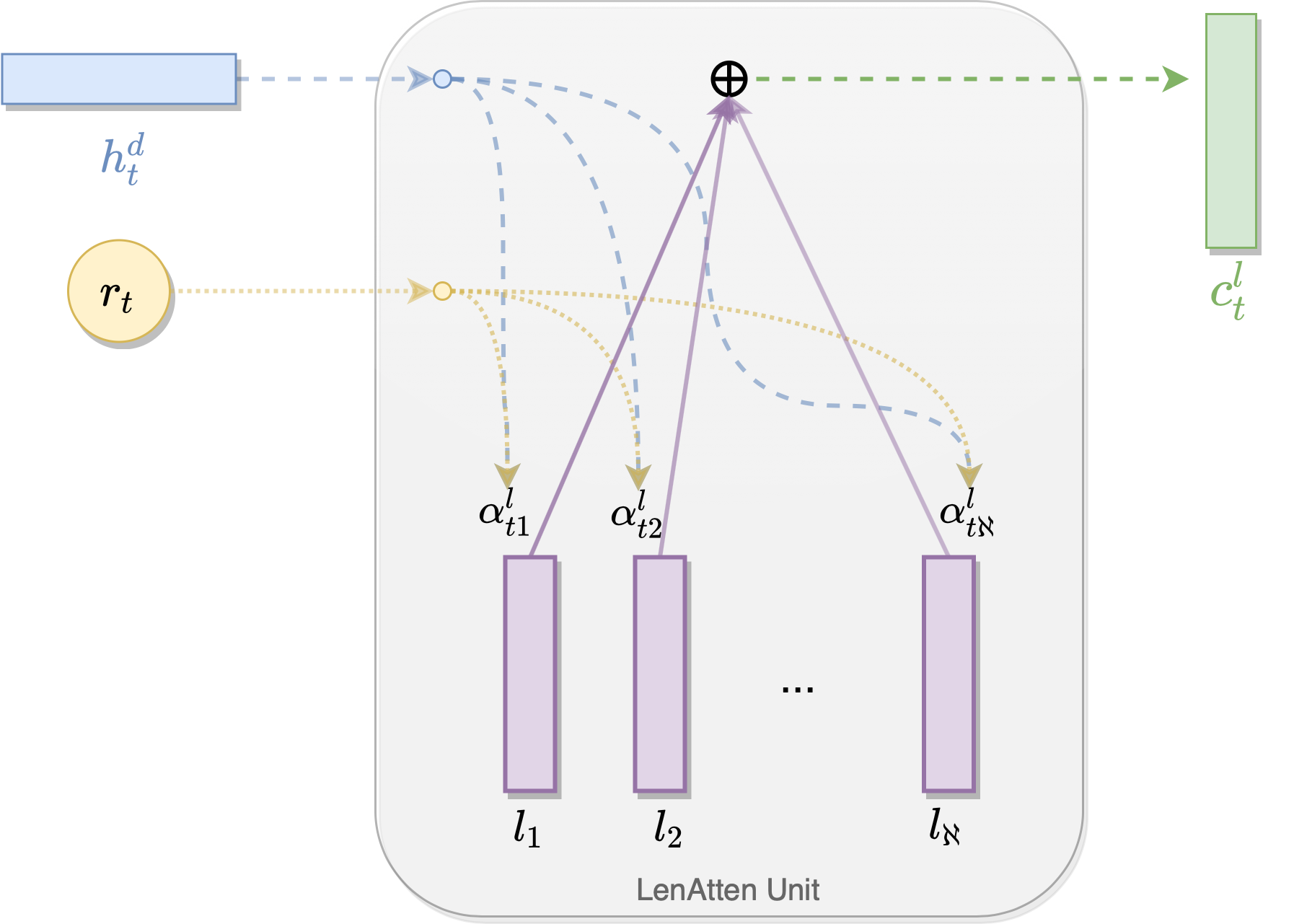}
  \caption{Illustration of the Length Attention Unit. Firstly, decoder hidden state (blue) and remaining length (yellow) are employed to compute the attention weights $a^l$. Then, the length context vector $c^l_t$ (green) is produced by calculating the weighted sum between attention weights and pre-defined length embeddings (purple). Better viewed in color.}
  \label{fig:lenatten-mechansim}
\end{figure}
The motivation of LenAtten is to separate length information from the input of the recurrent decoding unit and to exploit proper length information based on the stepwise remaining length. As shown in Figure \ref{fig:lenatten-mechansim}, at each decoding step, a length context vector is generated by calculating the weighted sum of a set of pre-defined embedding vectors $l_*$. Then, the length context vector is concatenated with the decoder hidden state and other attention vectors and fed to the input of the word prediction layer (details are shown in \S \ref{sec:methods_tb_compared}), so that summarizers can take the remaining length into account. The length context vector $c^l_t$ at $t$-th decoding step is defined as follows:
\begin{gather}
  c^l_t = \sum_{j=1}^{\aleph} \alpha^l_{tj}\, l_j\\
  \alpha^l_{tj} = \frac{exp(e^l_{tj})}{\sum_{k=1}^{\aleph} exp(e^l_{tk})}\\
  e^l_t = V^T_{l} \, tanh(W_{l} \, h^d_t + w_r \, r_t + b_{l}),
\end{gather} where $e^l_t \in \mathbb{R}^{{\aleph}\times 1}$, $\alpha^l_{tj}$ is the length attention score on the $j$-th length embedding at the $t$-th decoding step, $h^d_t$ is the decoder hidden state, and $V_{l}, W_{l}, w_r, b_{l}$ are learnable parameters. $r_t$ is a scalar representing the remaining length at the current decoding step and ${\aleph}$ is a hyperparameter indicating the number of pre-defined length embeddings. 

For length embeddings, we adopt the positional encoding proposed in \citet{DBLP:conf/nips/VaswaniSPUJGKP17}. We keep the embeddings fixed to remove the bias brought by the length distribution of data. The $j$-th length embedding $l_j$ is defined as follows:
\begin{equation}
  l_{j} =
  \begin{cases}
   0 & j = 1 \\
   PE(j-1) & otherwise,
  \end{cases}
\end{equation}
where $PE(\, \cdot \,)$ is the positional encoding.

At the $t$-th decoding step, the remaining length $r_t$ is updated by subtracting the length of the previously generated token. For the first decoding step, $r_1$ is initialized with desired output length. Following equations are used when $t>1$:
\begin{equation}
  r_{t} =
\begin{cases}
 0 & r_{t-1}- L(y_{t-1})  \leq 0 \\
 r_{t-1}- L(y_{t-1}) & otherwise
\end{cases}
\end{equation}
where $L(y_{t-1})$ returns the number of characters in the output word $y_{t-1}$.

\section{Experiment}

\subsection{Experimental Settings}
We evaluate LenAtten on the CNN/Daily Mail dataset \citep{see-etal-2017-get} to compare it with previous studies. In addition, we test LenAtten with short articles and summaries on the Annotated English Gigaword dataset \citep{rush-etal-2015-neural}. By default, all models are trained with maximum likelihood estimation (MLE) on a NVIDIA TITAN RTX GPU.\footnote{Detailed model configurations are provided in the Appendix.}
 
For evaluation metrics, we adopt the standard F1 score of ROUGE-1, ROUGE-2, and ROUGE-L \citep{lin-2004-rouge} to evaluate summary quality. For evaluating models' ability to control the output sequence length, we follow \citep{makino-etal-2019-global} to compute (1) character-level length variance $Var$ between reference summaries and generated summaries and (2) over-length ratio $\%over$, which measures how many of the generated summaries are longer than their reference summaries. The length variance $Var$ is computed as follows:
\begin{equation}
    Var = 0.001 * \frac{1}{n} \sum_{i=1}^n |len(y_i) - len(y_i')|^2
\end{equation} where $y_i$ is the reference summary, $y_i'$ is the generated summary, and $len(\cdot)$ returns the number of characters in the given summary. For the FLS task, the length variance $Var$ is expected to be zero as it indicates the lengths of output summaries are exactly the desired summary lengths.

\subsection{Methods to be compared}
\label{sec:methods_tb_compared}

We compare the proposed Length Attention unit with following methods:

\textsc{\textbf{LEAD-3}} extracts the first three sentences of source articles as the summary.

\textsc{\textbf{PG}} is the standard pointer-generator network proposed in \citet{see-etal-2017-get}.

\textsc{\textbf{MASS}} \citep{DBLP:conf/icml/SongTQLL19} is a sequence to sequence pre-trained model based on Transformer.

LenAtten is also compared with length controllable summarization methods. For a fair comparison, we choose methods that also aim at generating summaries with a preset number of characters in a word-by-word manner.

\textsc{\textbf{LE}} is the LenEmb method proposed in \citet{kikuchi-etal-2016-controlling}.

\textsc{\textbf{GOLC}} is a global optimization method introduced in \citet{makino-etal-2019-global}.

We apply LenAtten to three summarization models: 

\textsc{\textbf{S2S}} (RNN-based Seq2Seq Model) is a vanilla encoder-decoder summarizer. Specifically, we adopt a Bi-LSTM as the encoder and a unidirectional LSTM as the decoder. To integrate LenAtten, the length context vector $c^l_t$ is added to the input of the word prediction layer to produce the vocabulary distribution $P_{vocab}$:
\begin{equation}
    P_{vocab} = softmax\Big(W[h^d_t || c^l_t || y_{t-1}|| C] + b\Big)
\end{equation} where $W$, $b$ are learnable parameters, $h^d_t$ is the decoder hidden state, $y_{t-1}$ is the word embedding of the last generated token, ``$||$'' is the vector concatenation operator. $C$ is the last encoder hidden state, which is known as the fixed context vector.

\textsc{\textbf{Paulus}} (Copying Mechanism) follows the design of \citet {DBLP:conf/iclr/PaulusXS18}, which incorporates two attention modules and the copying mechanism into a Seq2seq summarizer. To integrate LenAtten, the vocabulary distribution $P_{vocab}$ is calculated using:
\begin{equation}
    P_{vocab} = softmax\Big(W[h^d_t || c^l_t || c^e_t || c^d_t] + b\Big)
\end{equation} where $c^e_t$ and $c^d_t$ are the context vectors generated from the encoder and decoder attention units.

\textsc{\textbf{Attention}} (Attention-based model) is implemented by removing copying mechanism from \textsc{\textbf{Paulus}}. For the above-mentioned three models, we remove the length context vector $c^l_t$ in the ablation study and keep other components unchanged.

\subsection{Experimental Results \footnote{For all the experiments, the number following \textsc{LA} (the LenAtten unit) represents the number of length embeddings (i.e. the value of $\aleph$).}}

\paragraph{Reference Summary Lengths}
\begin{table}[t]
\centering
\resizebox{0.48\textwidth}{!}{%
\begin{tabular}{lcccccc}
\hline
\multicolumn{6}{c}{CNN/DM}\\
Models    & R-1 F& R-2 F& R-L F& $Var\,(\downarrow)$ & $\%over$\\ \hline
Baseline  &&&     & &\\
\hdashline
\textsc{*LEAD-3} & 40.34& 17.70& 36.57&-&-\\
\textsc{$\dagger$MASS} & 41.38& 19.11& 38.42&-&-\\
\textsc{$\ddagger$PG} & 37.74& 15.78& 34.35& 19.35 & 58.35\\
\textsc{$\ddagger$PG + LE(mle)}& 37.45& 15.31& 34.28& 4.5 & 19.11\\
\textsc{$\ddagger$PG + LE(golc)}   & 38.27& 16.22& 34.99& 5.13  & 6.70\\
\textsc{S2S}    & 19.38& 3.58 & 14.35 & 89.74 & 7.00 \\
\textsc{Attention}  & 34.32& 13.76& 28.92 & 48.99 & 18.11\\
\textsc{Paulus}     & 38.10& 16.42& 33.17 & 19.91& 37.14\\
\hdashline    \multicolumn{6}{l}{with \textbf{Length Attention ($\aleph = 2$)}}  \\
\hdashline    \textbf{\textbf{S2S}}    & 21.09& 3.93 & 16.79& \textbf{0.0069} & 30.67\\
\textsc{\textbf{Attention}}& 36.53& 14.21& 32.63 & 0.0075& 38.72\\
\textsc{\textbf{Paulus}} & \textbf{39.82} & \textbf{17.31} & \textbf{36.20}& 0.0070 & 57.75\\ \hline
\end{tabular}%
}

\resizebox{0.48\textwidth}{!}{%
\begin{tabular}{lcccccc}
\hline
\multicolumn{6}{c}{AEG}\\
Models    & R-1 F& R-2 F& R-L F & $Var\,(\downarrow)$ & $\%over$  \\ \hline
Baseline  &&&     & &  \\
\hdashline
\textsc{S2S} & 36.99& 16.03& 33.01& 0.3902& 30.12\\
\textsc{Attention}  & 42.55& 21.54& 38.72& 0.3285& 25.89\\
\textsc{Paulus} & 43.84& \textbf{22.80} & 40.12& 0.3058& 16.01\\
\textsc{Paulus+LE}& 40.02&17.31&36.99&0.0500&0.4\\
\hdashline    \multicolumn{6}{l}{with \textbf{Length Attention ($\aleph = 2$)}}  \\
\hdashline    \textbf{\textbf{S2S}}  & 38.26& 16.24& 35.11& 0.0043& 35.45 \\
\textsc{\textbf{Attention}}& 43.15& 21.51& 40.32& 0.0044& 50.15 \\
\textsc{\textbf{Paulus}} & \textbf{43.92}\ & \textbf{22.80} & \textbf{41.16}& \textbf{0.0042} & 37.75\\ \hline
\end{tabular}%
  }
\caption{Results on CNN/DM and AEG dataset. If not specified in the parentheses, the training objective function is MLE by default. Results retrieved from: * \citet{see-etal-2017-get}; $\dagger$ \citet{xu-etal-2020-self}; $\ddagger$ \citet{makino-etal-2019-global}.}
\label{tab:gt-results}
\end{table}
\label{sec:experiment-gtsummarylength}
In this experiment, we evaluate our model by comparing it with previous works. The desired length is set as the number of characters in corresponding reference summaries. Table \ref{tab:gt-results} shows that LenAtten has superior length controllability and higher ROUGE scores on both datasets. Specifically, the length variance ($Var$) of LenAtten is 732 times better than the best-performing length controllable method \textsc{PG+LE(golc)} in the CNN/DM dataset. Besides, adding LenAtten can boost ROUGE scores by 1-3 points. We believe the improvement in the ROUGE scores comes from the introduction of length information (i.e. the desired length information). The desired length information can be viewed as an inductive bias, which helps summarizers prefer some of the outputs over others. Under the same context, by conditioning on the desired output length, summarizers may prefer candidate summaries with output lengths similar to the desired length. Thus, summarizers can learn a better alignment with the reference summaries during training and outputs summaries with higher ROUGE scores in inference.

In addition, previous length controllable methods control the output lengths at the cost of damaging the ROUGE scores. The ROUGE scores of \textsc{PG} and \textsc{Paulus} drop after adding LenEmb (i.e. \textsc{PG + LE(mle)} and \textsc{Paulus + LE}). In comparison, LenAtten not only performs better at reducing the length variance $Var$ but also significantly improves ROUGE scores. This suggests that LenAtten can break the trade-off between summary quality and length controllability.

After integrating with LenAtten, the $\%over$ ratio of summarizers rises. This suggests that more of the generated summaries ended up being longer than the references. We believe this is because when the remaining length is small (e.g. 4 characters) but not 0, instead of stopping the generation process, summarizers with LA tend to generate more tokens to meet the length requirement. Since summarizers output a word at each inference step, they may select a word that's longer than the remaining length. Thus, the generated summaries may end up being longer than the references.

\paragraph{Perplexity}
\begin{table}[t]
\centering
\resizebox{0.48\textwidth}{!}{%
\begin{tabular}{lcccccc}
\hline
Models & & & CNN/DM &&& AEG \\ \hline
\textsc{S2S} & & &  5.450 && &  3.624 \\
\textsc{S2S}+LA2 & & &  \textbf{5.360} && &  \textbf{3.393} \\
\hdashline
\textsc{Attention} & & &  4.317 && &  3.279 \\
\textsc{Attention}+LA2 & & &  \textbf{4.278} && &  \textbf{3.074} \\
\hdashline
\textsc{Paulus} & & &  3.478 && &  3.085 \\
\textsc{Paulus}+LA2 & & &  \textbf{3.391} && &  \textbf{2.899} \\ \hline
\end{tabular}%
}
\caption{Test perplexity of models on the CNN/DM dataset and the AEG dataset.}
\label{tab:perplexity}
\end{table}
To figure out how LenAtten affects the performance of the language model, we examine the log-perplexity of models on the test sets. Perplexity is a commonly-used metric for evaluating language models. A lower perplexity score indicates better language model performance. In this experiment, the desired length is set to the reference summary length. As shown in Table \ref{tab:perplexity}, after adding LenAtten, log-perplexity drops consistently on both datasets. This suggests that the adding of LenAtten can boost language model performance.
\paragraph{Various Preset Lengths}
\label{sec:experiment-fixedlength}

In this experiment, we test the generalization ability of LenAtten under various desired lengths. For the AEG dataset, most reference summaries contain 30-75 characters, and few of them are more than 100 characters. For the CNN/DM dataset, most reference summaries are 100-750 characters. Thus, the desired length is set as 30, 50, 75, 100, and 120 for the AEG dataset and 100, 200, 400, 800, and 1600 for the CNN/DM dataset. We add the LenAtten unit to \textsc{Paulus} and exploit full reference summaries to get ROUGE scores.

As shown in Table \ref{tab:fl-results}, on the AEG dataset, for frequently appeared lengths (30, 50, 75), and lengths that are exceptionally long (100, 120), LenAtten demonstrates great length controllability along with good ROUGE scores. Same conclusions can be drawn on the CNN/DM dataset.
This shows that LenAtten has great generalization ability under various desired lengths.

\begin{table}[t]
  \centering 
  \small
  \resizebox{0.48\textwidth}{!}{%
  \begin{tabular}{cccccc}
  \hline
  AEG&30&50&75&100&120 \\ \hline
  R-1 F & 37.89 & 42.67 & 41.13 & 39.03 & 37.45\\
  R-2 F & 18.87 & 21.01 & 18.99 & 17.16 & 16.00   \\
  R-L F & 33.64 & 38.76 & 35.27 & 31.95 & 29.84\\
  $Var$ & 0.0027 & 0.0024 & 0.0026 & 0.0030 & 0.0040 \\ \hline
  \end{tabular}
  }
  \\
  \small
  \resizebox{0.48\textwidth}{!}{%
  \begin{tabular}{cccccc}
  \hline
  {\tiny CNN/DM}&100&200&400&800&1600 \\ \hline
  R-1 F & 30.88 & 37.90 & 39.52 & 37.25 & 34.23\\
  R-2 F & 13.31 & 16.26 & 16.58 & 15.08 & 13.50   \\
  R-L F & 23.17 & 32.95 & 33.99 & 29.34 & 24.54\\
  $Var$ & 0.0067 & 0.0063 & 0.0058 & 0.0054 & 0.0051 \\ \hline
  \end{tabular}%
  }
  \caption{ROUGE scores and Length Variance $Var$ of \textsc{Paulus+LA2} under different desired lengths.}
  \label{tab:fl-results}
\end{table}

\begin{figure}[t]
  \centering
  \includegraphics[width=\linewidth]{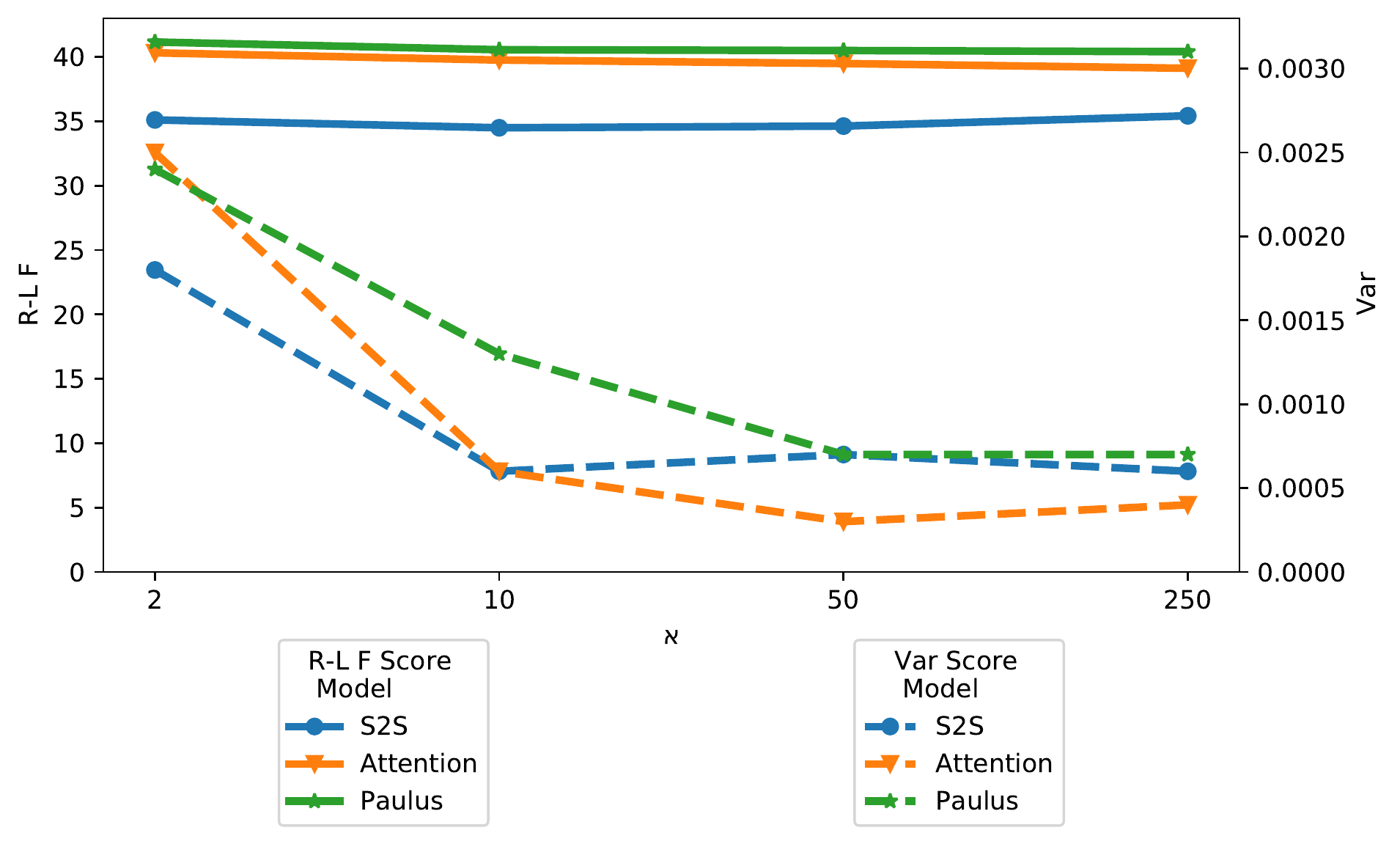}
  \caption{Examining hyperparameter $\aleph$ on the AEG dataset. ROUGE-L F1 scores and Length Variance $Var$ of different models under different $\aleph$ are shown $(\aleph = 2, 10, 50, 250)$.}
  \label{fig:aleph_search}
\end{figure}

\paragraph{Exploring Hyperparameter $\aleph$}
\label{sec:aleph-explore}
In this experiment, we analyze how different $\aleph$ (the number of pre-defined length embeddings) affect the performance of LenAtten on the AEG dataset. Desired lengths are set to the lengths of reference summaries. Figure \ref{fig:aleph_search} shows the length controllability becomes better as the increase of $\aleph$, with no harm to the ROUGE-L scores.

\section{Conclusions}
In this paper, we present a novel length controlling unit, LenAtten, to help summarization models generate quality summaries with a preset number of characters. On the examined datasets, LenAtten outperforms length controllable summarization baselines steadily in terms of length controllability and demonstrates great generalization ability. LenAtten also breaks the trade-off between length controllability and summary quality. To our knowledge, in the task of generating summaries with target lengths, LenAtten is the new state-of-the-art on the CNN Daily Mail dataset.

\section*{Acknowledgment}
We thank friends and colleagues at the UIC-AISIG as well as Minyong Li for their assistance with the study. We would like to thank anonymous reviewers for their helpful comments. This work is supported by the BNU-HKBU United International College research grant.

\bibliographystyle{acl_natbib}
\bibliography{lenatten}

\begin{thebibliography}{20}
\expandafter\ifx\csname natexlab\endcsname\relax\def\natexlab#1{#1}\fi

\bibitem[{Chopra et~al.(2016)Chopra, Auli, and
  Rush}]{chopra-etal-2016-abstractive}
Sumit Chopra, Michael Auli, and Alexander~M. Rush. 2016.
\newblock \href {https://doi.org/10.18653/v1/N16-1012} {Abstractive sentence
  summarization with attentive recurrent neural networks}.
\newblock In \emph{Proceedings of the 2016 Conference of the North {A}merican
  Chapter of the Association for Computational Linguistics: Human Language
  Technologies}, pages 93--98, San Diego, California. Association for
  Computational Linguistics.

\bibitem[{Dorr et~al.(2003)Dorr, Zajic, and Schwartz}]{dorr-etal-2003-hedge}
Bonnie Dorr, David Zajic, and Richard Schwartz. 2003.
\newblock \href {https://www.aclweb.org/anthology/W03-0501} {Hedge trimmer: A
  parse-and-trim approach to headline generation}.
\newblock In \emph{Proceedings of the {HLT}-{NAACL} 03 Text Summarization
  Workshop}, pages 1--8.

\bibitem[{Fan et~al.(2018)Fan, Grangier, and Auli}]{fan-etal-2018-controllable}
Angela Fan, David Grangier, and Michael Auli. 2018.
\newblock \href {https://doi.org/10.18653/v1/W18-2706} {Controllable
  abstractive summarization}.
\newblock In \emph{Proceedings of the 2nd Workshop on Neural Machine
  Translation and Generation}, pages 45--54, Melbourne, Australia. Association
  for Computational Linguistics.

\bibitem[{Gu et~al.(2016)Gu, Lu, Li, and Li}]{gu-etal-2016-incorporating}
Jiatao Gu, Zhengdong Lu, Hang Li, and Victor~O.K. Li. 2016.
\newblock \href {https://doi.org/10.18653/v1/P16-1154} {Incorporating copying
  mechanism in sequence-to-sequence learning}.
\newblock In \emph{Proceedings of the 54th Annual Meeting of the Association
  for Computational Linguistics (Volume 1: Long Papers)}, pages 1631--1640,
  Berlin, Germany. Association for Computational Linguistics.

\bibitem[{Kikuchi et~al.(2016)Kikuchi, Neubig, Sasano, Takamura, and
  Okumura}]{kikuchi-etal-2016-controlling}
Yuta Kikuchi, Graham Neubig, Ryohei Sasano, Hiroya Takamura, and Manabu
  Okumura. 2016.
\newblock \href {https://doi.org/10.18653/v1/D16-1140} {Controlling output
  length in neural encoder-decoders}.
\newblock In \emph{Proceedings of the 2016 Conference on Empirical Methods in
  Natural Language Processing}, pages 1328--1338, Austin, Texas. Association
  for Computational Linguistics.

\bibitem[{Lin(2004)}]{lin-2004-rouge}
Chin-Yew Lin. 2004.
\newblock \href {https://www.aclweb.org/anthology/W04-1013} {{ROUGE}: A package
  for automatic evaluation of summaries}.
\newblock In \emph{Text Summarization Branches Out}, pages 74--81, Barcelona,
  Spain. Association for Computational Linguistics.

\bibitem[{Liu and Lapata(2019)}]{liu-lapata-2019-text}
Yang Liu and Mirella Lapata. 2019.
\newblock \href {https://doi.org/10.18653/v1/D19-1387} {Text summarization with
  pretrained encoders}.
\newblock In \emph{Proceedings of the 2019 Conference on Empirical Methods in
  Natural Language Processing and the 9th International Joint Conference on
  Natural Language Processing (EMNLP-IJCNLP)}, pages 3730--3740, Hong Kong,
  China. Association for Computational Linguistics.

\bibitem[{Liu et~al.(2018)Liu, Luo, and Zhu}]{liu-etal-2018-controlling}
Yizhu Liu, Zhiyi Luo, and Kenny Zhu. 2018.
\newblock \href {https://doi.org/10.18653/v1/D18-1444} {Controlling length in
  abstractive summarization using a convolutional neural network}.
\newblock In \emph{Proceedings of the 2018 Conference on Empirical Methods in
  Natural Language Processing}, pages 4110--4119, Brussels, Belgium.
  Association for Computational Linguistics.

\bibitem[{Makino et~al.(2019)Makino, Iwakura, Takamura, and
  Okumura}]{makino-etal-2019-global}
Takuya Makino, Tomoya Iwakura, Hiroya Takamura, and Manabu Okumura. 2019.
\newblock \href {https://doi.org/10.18653/v1/P19-1099} {Global optimization
  under length constraint for neural text summarization}.
\newblock In \emph{Proceedings of the 57th Annual Meeting of the Association
  for Computational Linguistics}, pages 1039--1048, Florence, Italy.
  Association for Computational Linguistics.

\bibitem[{Nallapati et~al.(2017)Nallapati, Zhai, and
  Zhou}]{nallapati2016summarunner-AAAI1714636}
Ramesh Nallapati, Feifei Zhai, and Bowen Zhou. 2017.
\newblock \href
  {https://aaai.org/ocs/index.php/AAAI/AAAI17/paper/view/14636/14080}
  {Summarunner: A recurrent neural network based sequence model for extractive
  summarization of documents}.
\newblock In \emph{Proceedings of the AAAI Conference on Artificial
  Intelligence, 31(1)}.

\bibitem[{Nallapati et~al.(2016)Nallapati, Zhou, dos Santos,
  G{\"{u}}l{\c{c}}ehre, and Xiang}]{DBLP:conf/conll/NallapatiZSGX16}
Ramesh Nallapati, Bowen Zhou, C{\'{\i}}cero~Nogueira dos Santos, {\c{C}}aglar
  G{\"{u}}l{\c{c}}ehre, and Bing Xiang. 2016.
\newblock \href {https://www.aclweb.org/anthology/K16-1028/} {Abstractive text
  summarization using sequence-to-sequence rnns and beyond}.
\newblock In \emph{Proceedings of the 20th {SIGNLL} Conference on Computational
  Natural Language Learning, CoNLL 2016, Berlin, Germany, August 11-12, 2016},
  pages 280--290. {ACL}.

\bibitem[{Paulus et~al.(2018)Paulus, Xiong, and
  Socher}]{DBLP:conf/iclr/PaulusXS18}
Romain Paulus, Caiming Xiong, and Richard Socher. 2018.
\newblock \href {https://openreview.net/forum?id=HkAClQgA-} {A deep reinforced
  model for abstractive summarization}.
\newblock In \emph{6th International Conference on Learning Representations,
  {ICLR} 2018, Vancouver, BC, Canada, April 30 - May 3, 2018, Conference Track
  Proceedings}. OpenReview.net.

\bibitem[{Rush et~al.(2015)Rush, Chopra, and Weston}]{rush-etal-2015-neural}
Alexander~M. Rush, Sumit Chopra, and Jason Weston. 2015.
\newblock \href {https://doi.org/10.18653/v1/D15-1044} {A neural attention
  model for abstractive sentence summarization}.
\newblock In \emph{Proceedings of the 2015 Conference on Empirical Methods in
  Natural Language Processing}, pages 379--389, Lisbon, Portugal. Association
  for Computational Linguistics.

\bibitem[{See et~al.(2017)See, Liu, and Manning}]{see-etal-2017-get}
Abigail See, Peter~J. Liu, and Christopher~D. Manning. 2017.
\newblock \href {https://doi.org/10.18653/v1/P17-1099} {Get to the point:
  Summarization with pointer-generator networks}.
\newblock In \emph{Proceedings of the 55th Annual Meeting of the Association
  for Computational Linguistics (Volume 1: Long Papers)}, pages 1073--1083,
  Vancouver, Canada. Association for Computational Linguistics.

\bibitem[{Song et~al.(2019)Song, Tan, Qin, Lu, and
  Liu}]{DBLP:conf/icml/SongTQLL19}
Kaitao Song, Xu~Tan, Tao Qin, Jianfeng Lu, and Tie{-}Yan Liu. 2019.
\newblock \href {http://proceedings.mlr.press/v97/song19d.html} {{MASS:} masked
  sequence to sequence pre-training for language generation}.
\newblock In \emph{Proceedings of the 36th International Conference on Machine
  Learning, {ICML} 2019, 9-15 June 2019, Long Beach, California, {USA}},
  volume~97 of \emph{Proceedings of Machine Learning Research}, pages
  5926--5936. {PMLR}.

\bibitem[{Takase and Okazaki(2019)}]{takase-okazaki-2019-positional}
Sho Takase and Naoaki Okazaki. 2019.
\newblock \href {https://doi.org/10.18653/v1/N19-1401} {Positional encoding to
  control output sequence length}.
\newblock In \emph{Proceedings of the 2019 Conference of the North {A}merican
  Chapter of the Association for Computational Linguistics: Human Language
  Technologies, Volume 1 (Long and Short Papers)}, pages 3999--4004,
  Minneapolis, Minnesota. Association for Computational Linguistics.

\bibitem[{Vaswani et~al.(2017)Vaswani, Shazeer, Parmar, Uszkoreit, Jones,
  Gomez, Kaiser, and Polosukhin}]{DBLP:conf/nips/VaswaniSPUJGKP17}
Ashish Vaswani, Noam Shazeer, Niki Parmar, Jakob Uszkoreit, Llion Jones,
  Aidan~N. Gomez, Lukasz Kaiser, and Illia Polosukhin. 2017.
\newblock \href {http://papers.nips.cc/paper/7181-attention-is-all-you-need}
  {Attention is all you need}.
\newblock In \emph{Advances in Neural Information Processing Systems 30: Annual
  Conference on Neural Information Processing Systems 2017, 4-9 December 2017,
  Long Beach, CA, {USA}}, pages 5998--6008.

\bibitem[{Xu et~al.(2020)Xu, Li, Yuan, Wu, He, and Zhou}]{xu-etal-2020-self}
Song Xu, Haoran Li, Peng Yuan, Youzheng Wu, Xiaodong He, and Bowen Zhou. 2020.
\newblock \href {https://doi.org/10.18653/v1/2020.acl-main.125} {Self-attention
  guided copy mechanism for abstractive summarization}.
\newblock In \emph{Proceedings of the 58th Annual Meeting of the Association
  for Computational Linguistics}, pages 1355--1362, Online. Association for
  Computational Linguistics.

\bibitem[{Zhang et~al.(2019)Zhang, Kishore, Wu, Weinberger, and
  Artzi}]{bert-score}
Tianyi Zhang, Varsha Kishore, Felix Wu, Kilian~Q. Weinberger, and Yoav Artzi.
  2019.
\newblock \href {http://arxiv.org/abs/1904.09675} {Bertscore: Evaluating text
  generation with {BERT}}.
\newblock \emph{CoRR}, abs/1904.09675.

\bibitem[{Zhong et~al.(2020)Zhong, Liu, Chen, Wang, Qiu, and
  Huang}]{zhong-etal-2020-extractive}
Ming Zhong, Pengfei Liu, Yiran Chen, Danqing Wang, Xipeng Qiu, and Xuanjing
  Huang. 2020.
\newblock \href {https://doi.org/10.18653/v1/2020.acl-main.552} {Extractive
  summarization as text matching}.
\newblock In \emph{Proceedings of the 58th Annual Meeting of the Association
  for Computational Linguistics}, pages 6197--6208, Online. Association for
  Computational Linguistics.

\end{thebibliography}

\clearpage
\appendix
\section{Appendix}
\subsection{Experimental settings}
\paragraph{Model Configuration}
In our experiments, the dimension of LSTM hidden state is set as 256, and the vocabulary size is 100,000. Word embeddings are fixed 300-dimensional GloVe vectors\footnote{http://nlp.stanford.edu/data/glove.6B.zip}. If a word is not covered in the GloVe, a random 300-dimensional vector is used. During training, Adam optimizer is applied with $\ \beta_1=0.9, \ \beta_2 = 0.999, \ \epsilon = 10^{-8}$ and learning rate $\alpha = 0.001$. Besides, we set a $25\%$ probability of choosing the previously generated token instead of the ground-truth token as $y_{t-1}$ to reduce exposure bias. At test time, summaries are produced using beam search with beam size 4. We use a fully python implemented library\footnote{https://github.com/pltrdy/rouge} to obtain the ROUGE score.

\begin{figure}[!htp]
\centering
\includegraphics[width=\linewidth]{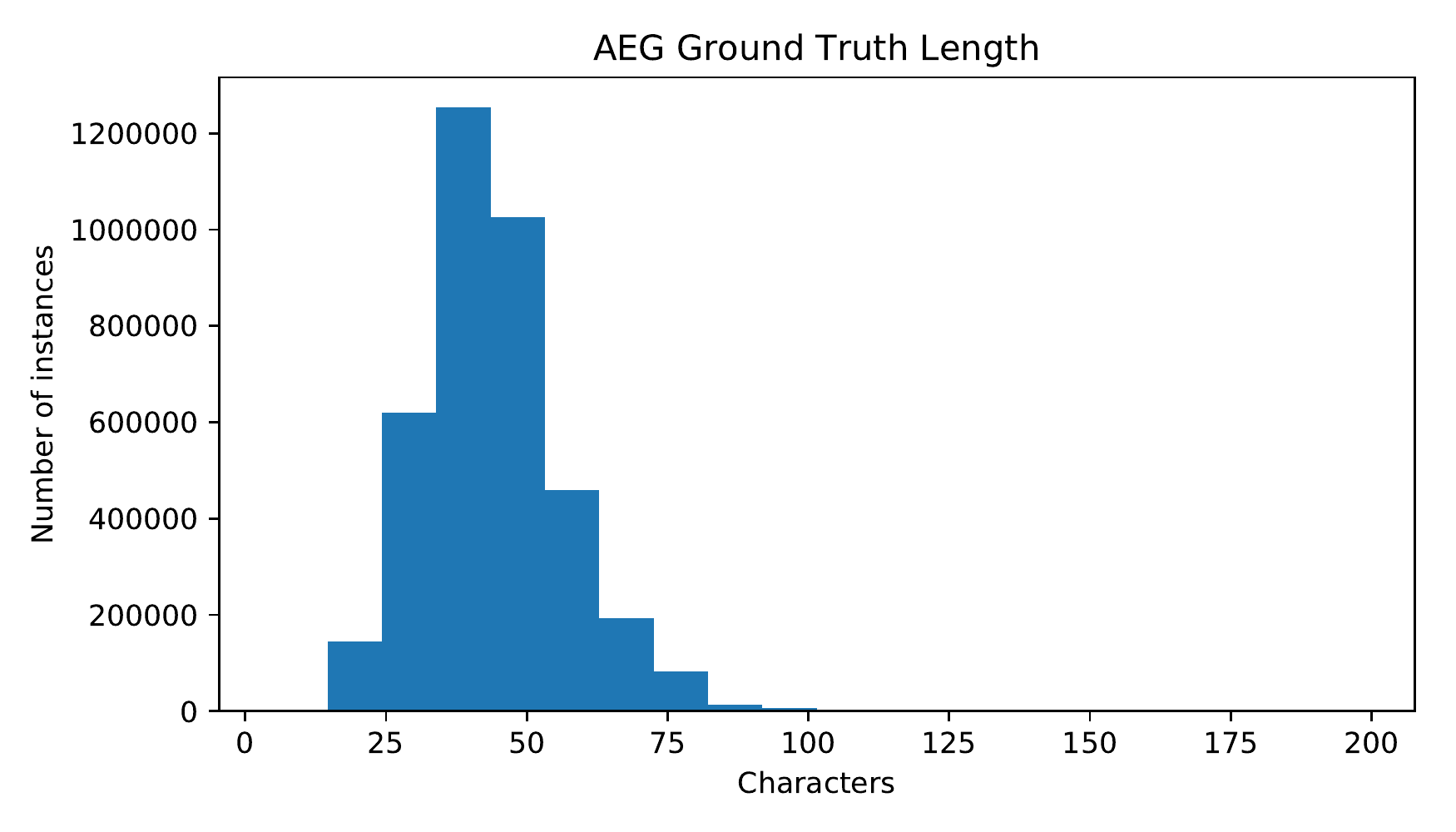}
\caption{Length distribution of reference summaries on the Annotated English Gigaword dataset. Summaries with 30 to 75 characters cover the majority cases.}
\label{fig:aeg-distribution}
\end{figure}

\begin{figure}[!htp]
\centering
\includegraphics[width=\linewidth]{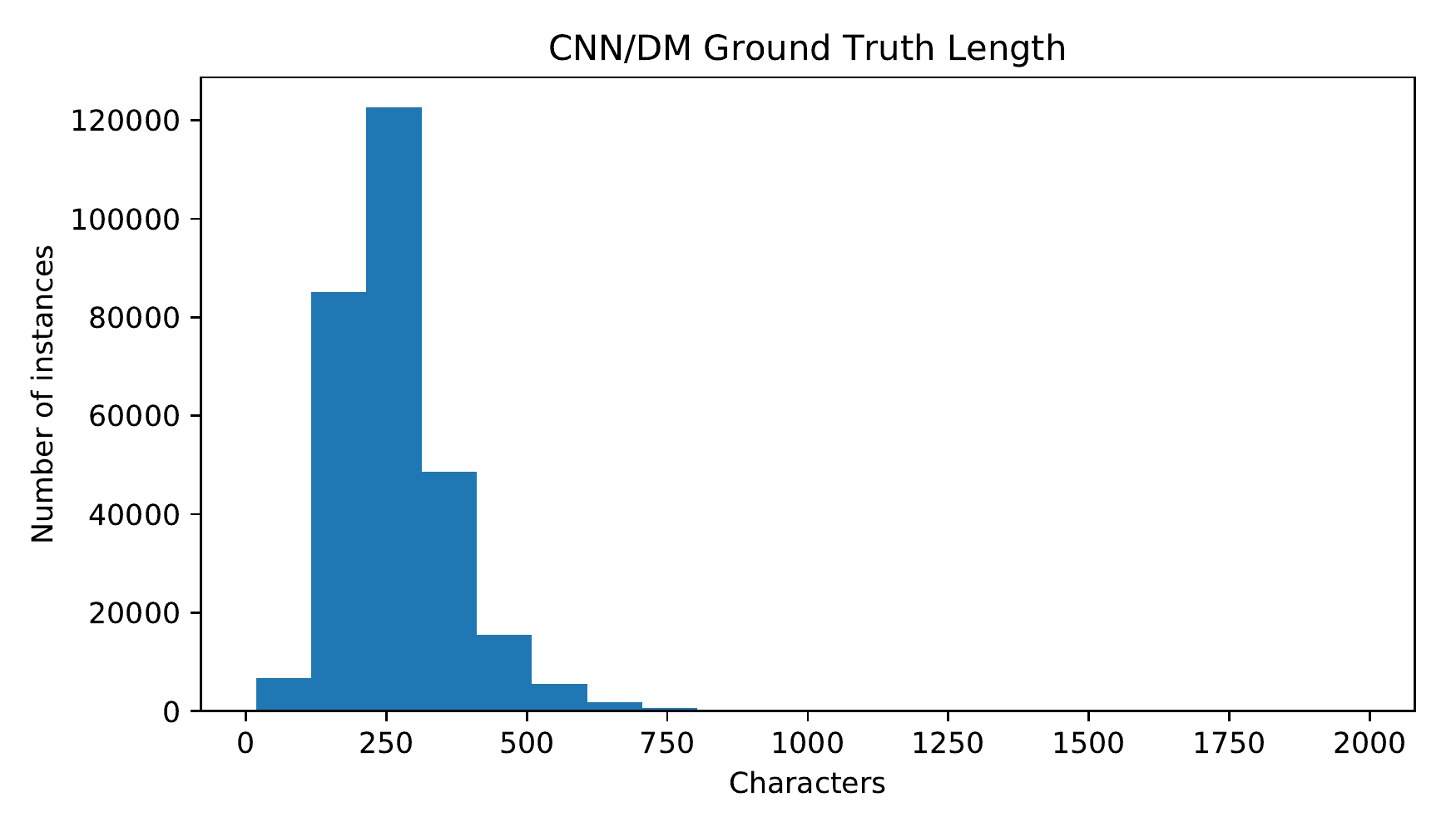}
\caption{Length distribution of reference summaries on the CNN/Daily Mail dataset. Summaries exceed 2000 characters are ignored, since they only cover 0.009\% of the dataset.}
\label{fig:cd-distribution}
\end{figure}

\paragraph{Dataset Distribution}
We plot the length distribution of reference summaries in the AEG (Figure \ref{fig:aeg-distribution}) and the CNN/DM (Figure \ref{fig:cd-distribution}) dataset.

\subsection{Additional Experiments}
\begin{figure}[!htb]
  \centering
  \includegraphics[width=\linewidth]{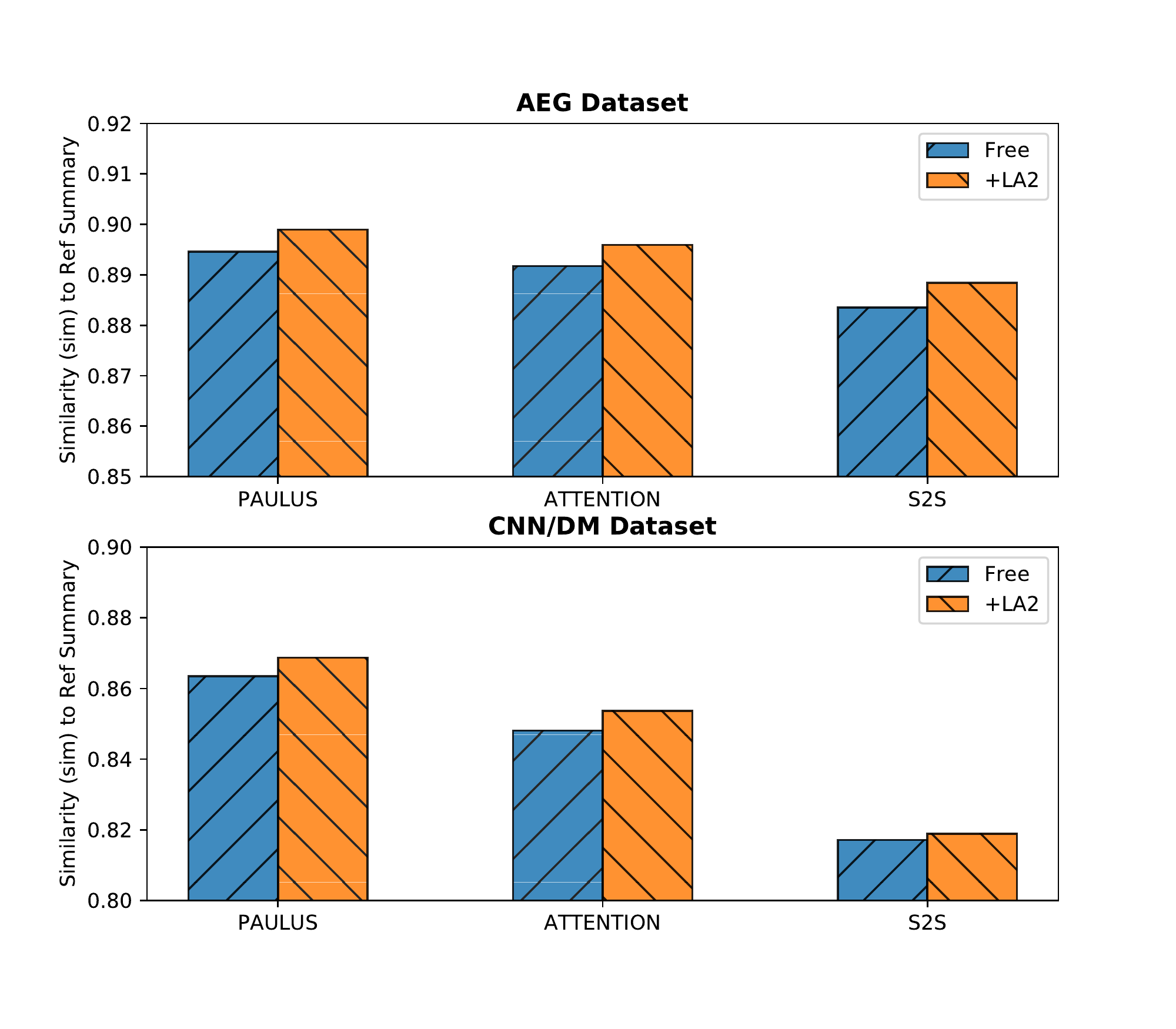}
  \caption{Semantic similarity between model outputs and reference summaries. Desired length is set as reference summary length.}
  \label{fig:sim_result}
\end{figure}

\paragraph{Semantic Similarity}
Another automatic evaluation metric \textit{BertScore} \cite{bert-score} recall score is used to measure the semantic similarity between system outputs and reference summaries. As shown in Figure \ref{fig:sim_result}, models with Length Attention module (\textsc{LA2}) outperform baselines (\textsc{Free}) on both datasets. 

\paragraph{Human Evaluation}
\label{sec:experiment-human}
Correctness (CORR), completeness (COMP), and fluency (FLUE) of system outputs are assessed through 2 human evaluations. We randomly select 10 samples from each dataset. 30 skilled English speakers are presented with the original article and two summaries. One of the summaries is from the model without LenAtten, and the other one is from the same model plus LenAtten (e.g., \textsc{S2S} and \textsc{S2S+LA2}). The evaluation process is well-designed to prevent participants from knowing the source of the presented summaries. Models without LenAtten generate summaries without length restriction, and models with LenAtten are required to output summaries with desired lengths. There are 467 feedbacks collected for the first experiment and 160 for the second.

In the first experiment (Table \ref{tab:human-evaluation-1}), participants are asked to choose a better one from two given summaries. The desired length is set as the length of the reference summary.

In the second experiment (Table \ref{tab:human-evaluation-2}), we only examine \textsc{Paulus} and \textsc{Paulus+LA2}. The desired length is set as (30, 50, 70) on the AEG dataset and (150, 250, 350) on the CNN/DM dataset. Participants need to rate each summary from 0 to 5. In order to guarantee the accuracy and credibility of results, each article is presented once to each participant.
\begin{table}[!tp]
  \centering
  \footnotesize
  \begin{tabular}{lccc}
  \hline
& COMP & CORR & FLUE \\ \hline
  \textbf{AEG}                   &              &             &         \\\hdashline
  \textsc{S2S} (Free)        & 27.1\%        & 44.4\%       & 42.9\%   \\
  \textsc{S2S+LA2} (GT)      & \textbf{72.9\%}        & \textbf{55.6\%}       & \textbf{57.1\%}   \\\hdashline
  \textsc{Attention} (Free)     & 33.3\%        & 44.2\%       & \textbf{56.9\%}   \\
  \textsc{Attention+LA2} (GT) & \textbf{66.7\%}        & \textbf{55.8\%}       & 43.1\%   \\\hdashline
  \textsc{Paulus} (Free)         & 47.2\%        & \textbf{53.3\%}       & \textbf{61.4\%}   \\
  \textsc{Paulus+LA2} (GT)     & \textbf{52.8\%}        & 46.7\%       & 38.6\%   \\ \hline
  \textbf{CNN/DM}                &              &             &         \\\hdashline
  \textsc{S2S} (Free)        & 22.5\%        & 22.9\%       & \textbf{53.8\%}   \\
  \textsc{S2S+LA2} (GT)      & \textbf{77.5\%}        & \textbf{77.1\%}       & 46.2\%   \\\hdashline
  \textsc{Attention} (Free)     & 21.9\%        & 27.5\%       & 36.7\%   \\
  \textsc{Attention}+LA2 (GT) & \textbf{78.1\%}        & \textbf{72.5\%}       & \textbf{63.3\%}   \\\hdashline
  \textsc{Paulus} (Free)         & \textbf{53\%}        & 47.4\%       & 48.1\%   \\
  \textsc{Paulus+LA2} (GT)     & 47\%         & \textbf{52.6\%}       & \textbf{51.9\%}   \\ \hline
  \end{tabular}
  \caption{Results of the first Human Evaluation. ``(Free)'': model generates summaries freely. ``(GT)'': model generates summaries with the desired length set as the length of the reference summary.}
  \label{tab:human-evaluation-1}
  \end{table}

\begin{table}[t]
\centering
\small
\resizebox{0.48\textwidth}{!}{%
\begin{tabular}{lccc}
\hline
                              & COMP & CORR & FLUE \\ \hline
\textbf{AEG}                           &              &             &         \\\hdashline
\textsc{Paulus} (Free, avg\_len=57.46) & 3.328        & 3.407       & 3.605   \\
\textsc{Paulus+LA2} (30)             & 3.000            & 3.250        & 3.392   \\
\textsc{Paulus+LA2} (50)         & 3.750         & 3.500         & \textbf{3.687}   \\
\textsc{Paulus+LA2} (70)             & \textbf{3.875}        & \textbf{3.750}        & 3.562   \\\hline
\textbf{CNN/DM}                        &              &             &         \\\hdashline
\textsc{Paulus} (Free, avg\_len=285.5) & 3.250         & 3.345       & 3.273   \\
\textsc{Paulus+LA2} (150)             & 3.125        & 3.375       & 3.166   \\
\textsc{Paulus+LA2} (250)        & 3.451        & \textbf{3.483}       & \textbf{3.419}   \\
\textsc{Paulus+LA2} (350)            & \textbf{3.827}        & 3.482       & 3.172   \\ \hline
\end{tabular}%
}
\caption{Results of the second human evaluation. ``(Free, avg\_len)'': model generates summaries freely. The average length of the generated summaries is also listed.}
\label{tab:human-evaluation-2}
\end{table}

As shown in Table \ref{tab:human-evaluation-1}, models with LenAtten have better completeness and correctness scores on both datasets, along with a few improvements on the fluency. In the second experiment, Table \ref{tab:human-evaluation-2} shows that (1) the completeness and correctness scores increase as the desired length increases. This trend is reasonable, since more information should be included in the final summary, as the summary length gets longer. This also suggests that, as the desired length gets longer, models with LenAtten can generate meaningful words instead of simply repeating one or two words. (2) When comparing the results of \textsc{Paulus} (Free, 57.46; 285.5) and \textsc{Paulus+LA2} (50; 250), \textsc{Paulus+LA2} outperforms the \textsc{Paulus} (Free) on all metrics. In other words, when the desired length gets smaller (but not too small), LenAtten can help summarization models to use concise words and phrases while maintaining summary quality.

\subsection{Output Examples}
\paragraph{Synonym substitution}
When examining generated summaries, we find adding LenAtten can make summarizers replace long/short words with synonyms to meet the length requirement. Examples are showcased in Table \ref{tab:generated-example-synonym}.

\begin{table}[t]
\centering
\resizebox{0.48\textwidth}{!}{%
\begin{tabular}{llcc}
\hline
\multicolumn{4}{l}{\textit{\textbf{Source document - A}}} \\ \hline
\multicolumn{4}{l}{\begin{tabular}{@{}l@{}}the indian union government thursday decided to \\increase \textcolor{red}{customs duty} on sugar to \#\# percent to curb\\ cheap imports of the commodity , said a senior finance\\ ministry official here .\end{tabular}} \\ \hline
\multicolumn{4}{l}{\textit{\textbf{Reference summary - A}}} \\ \hline
\multicolumn{4}{l}{india increases sugar \textcolor{red}{import duty}} \\ \hline
 & Summary & R-1 R & Diff \\ \hline
Model & \begin{tabular}{@{}l@{}}india to increase\\ \textcolor{red}{customs duty} on sugar\end{tabular} & 60.00 & 4 \\\hdashline
Model + LA2 & \begin{tabular}{@{}l@{}}india to increase\\ \textcolor{red}{tariffs} on sugar\end{tabular} & 40.00 & 0 \\ \hline
\end{tabular}%
}
\resizebox{0.48\textwidth}{!}{%
\begin{tabular}{llcc}
\hline
\multicolumn{4}{l}{\textit{\textbf{Source document - B}}} \\ \hline
\multicolumn{4}{l}{\begin{tabular}{@{}l@{}}defending champion \textbf{albert costa} of spain reached the\\ \textcolor{red}{last eight} in the french open here on monday , beating\\ local favorite \#\#nd seed arnaud clement of france in\\ straight sets .\end{tabular}} \\ \hline
\multicolumn{4}{l}{\textit{\textbf{Reference summary - B}}} \\ \hline
\multicolumn{4}{l}{\textbf{costa} enters \textcolor{red}{last eight} in french open} \\ \hline
 & Summary & R-1 R & Diff \\ \hline
Model & \begin{tabular}{@{}l@{}}\textbf{costa} into \textcolor{red}{last eight}\\ in french open\end{tabular} & 85.71 & -2 \\\hdashline
Model + LA2 & \begin{tabular}{@{}l@{}}\textbf{costa} into french open\\ \textcolor{red}{quarter-finals}\end{tabular} & 42.85 & -1 \\ \hline
\end{tabular}%
}
\caption{Synonym substitution is colored in red. \textbf{R-1 R}: ROUGE-1 Recall score. \textbf{Diff}: len(output) - len(reference).}
\label{tab:generated-example-synonym}
\end{table}

\end{document}